\documentclass[runningheads]{llncs}
\usepackage{graphicx}
\usepackage{amsmath,amssymb} 
\usepackage{color}

\usepackage{epsfig}
\usepackage{authblk}

\usepackage{epstopdf, booktabs}
\usepackage{enumitem}
\usepackage{subfigure}
\usepackage{array}
\usepackage{multirow}
\usepackage{xspace}
\usepackage{cite}
\usepackage{setspace}

\DeclareMathAlphabet{\mathpzc}{OT1}{pzc}{m}{it} 

\usepackage{amsfonts}

\makeatletter
\DeclareRobustCommand\onedot{\futurelet\@let@token\@onedot}
\def\@onedot{\ifx\@let@token.\else.\null\fi\xspace}

\newcolumntype{L}[1]{>{\raggedright\let\newline\\\arraybackslash}m{#1}}
\newcolumntype{C}[1]{>{\centering\let\newline\\\arraybackslash}m{#1}}
\newcolumntype{R}[1]{>{\raggedleft\let\newline\\\arraybackslash}m{#1}}

\def\eg{\emph{e.g}\onedot} 
\def\ie{\emph{i.e}\onedot}

\def\etal{\emph{et al}\onedot}

\def\@fnsymbol#1{\ensuremath{\ifcase#1\or *\or \dagger\or \ddagger\or
   \mathsection\or \mathparagraph\or \|\or **\or \dagger\dagger
   \or \ddagger\ddagger \else\@ctrerr\fi}}

\makeatother

\begin{document}
\mainmatter
\title{Show, Tell and Discriminate: Image Captioning by Self-retrieval with Partially Labeled Data} 
\titlerunning{Show, Tell and Discriminate}
\authorrunning{X. Liu. H. Li, J. Shao, D. Chen, X. Wang}
\author{Xihui Liu\inst{1} \and
Hongsheng Li\footnote{Hongsheng Li is the corresponding author.}\inst{1} \and
Jing Shao\inst{2} \and
Dapeng Chen\inst{1} \and
Xiaogang Wang\inst{1}}
\institute{$^1$The Chinese University of Hong Kong~~~~~~$^2$SenseTime Research\\
\email{\{xihui-liu@link.,hsli@ee.,dpchen@,xgwang@ee.\}cuhk.edu.hk\\shaojing@sensetime.com}}
\maketitle
\begin{abstract}
The aim of image captioning is to generate captions by machine to describe image contents.
Despite many efforts, generating discriminative captions for images remains non-trivial.
Most traditional approaches imitate the language structure patterns, thus tend to fall into a stereotype of replicating frequent phrases or sentences and neglect unique aspects of each image.
In this work, we propose an image captioning framework with a self-retrieval module as training guidance, which encourages generating discriminative captions. 
It brings unique advantages: (1) the self-retrieval guidance can act as a metric and an evaluator of caption discriminativeness to assure the quality of generated captions. 
(2) The correspondence between generated captions and images are naturally incorporated in the generation process without human annotations, and hence our approach could utilize a large amount of unlabeled images to boost captioning performance with no additional annotations.
We demonstrate the effectiveness of the proposed retrieval-guided method on COCO and Flickr$30$k captioning datasets, and show its superior captioning performance with more discriminative captions.

\keywords{image captioning, language and vision, text-image retrieval}
\end{abstract}

\section{Introduction}

\begin{figure}[t]
\centering
\includegraphics[width=1\linewidth]{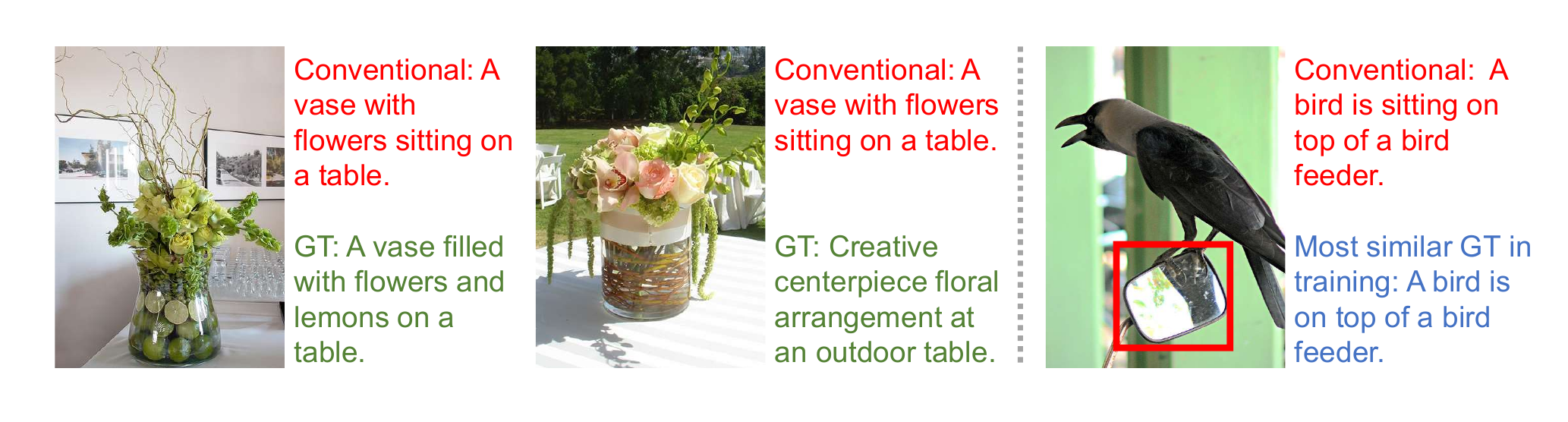}
\caption{Examples of generated captions by conventional captioning models. The generated captions are templated and generic.}
\label{fig1_current_captions}
\end{figure}

Image captioning, generating natural language description for a given image, is a crucial task that has drawn remarkable attention in the field of vision and language~\cite{vinyals2015show,xu2015show,yao2017boosting,chen2017sca,lu2017knowing,gan2017semantic,wang2017skeleton,anderson2017bottom,rennie2017self,li2018factorizable,li2017scene}. 
However, results by existing image captioning methods tend to be generic and templated. 
For example, in Fig.~\ref{fig1_current_captions}, although for humans there are non-neglectable differences between the first and second images, the captioning model gives identical ambiguous descriptions ``A vase with flowers sitting on a table", while the ground-truth captions contain details and clearly show the differences between those images. Moreover, about fifty percent of the captions generated by conventional captioning methods are exactly the same as ground-truth captions from the training set, indicating that the captioning models only learn a stereotype of sentences and phrases in the training set, and have limited ability of generating discriminative captions. The image on the right part of Fig.~\ref{fig1_current_captions} shows that although the bird is standing on a mirror, the captioning model generates the caption ``A bird is sitting on top of a bird feeder'', as a result of replicating patterns appeared in the training set.

Existing studies working on the aforementioned problems either used Generative Adversarial Networks (GAN) to generate human-like descriptions~\cite{dai2017towards,shetty2017speaking}, or focused on enlarging the diversity of generated captions~\cite{wang2017diverse,wang2016diverse,vijayakumar2016diverse}. Those methods improve the diversity of generated captions but sacrifice overall performance on standard evaluation criteria. Another work~\cite{vedantam2017context} generates discriminative captions for an image in context of other semantically similar images by an inference technique on both target images and distractor images, which cannot be applied to generic captioning where distractor images are not provided.
%

In this study, we wish to show that with the innovative model design, both the discriminativeness and fidelity can be effectively improved for caption generation.
It is achieved by involving a self-retrieval module to train a captioning module, motivated from two aspects: (1) the discriminativeness of a caption can be evaluated by how well it can distinguish its corresponding image from other images. This criterion can be introduced as a guidance for training, and thus encourages discriminative captions. (2) Image captioning and text-to-image retrieval can be viewed as dual tasks. Image captioning generates a description of a given image, while text-to-image retrieval retrieves back the image based on the generated caption.
Specifically, the model consists of a \textbf{Captioning Module} and a \textbf{Self-retrieval Module}. The captioning module generates captions based on given images, while the self-retrieval module conducts text-to-image retrieval, trying to retrieve corresponding images based on the generated captions. It acts as an evaluator to measure the quality of captions and encourages the model to generate discriminative captions. 
Since generating each word of a caption contains non-differentiable operations, we take the negative retrieval loss as self-retrieval reward and adopt REINFORCE algorithm to compute gradients.

Such retrieval-guided captioning framework can not only guarantee the discriminativeness of captions, but also readily obtain benefits from additional unlabeled images, since a caption naturally corresponds to the image it is generated from, and do not need laborious annotations. In detail, for unlabeled images, only self-retrieval module is used to calculate reward, while for labeled images, both the ground-truth captions and self-retrieval module are used to calculate reward and optimize the captioning model. Mining moderately hard negative samples from unlabeled data further boost both the fidelity and discriminativeness of image captioning.

We test our approach on two image captioning datasets, COCO~\cite{chen2015microsoft} and Flickr$30$k~\cite{young2014image}, in fully-supervised and semi-supervised settings. Our approach achieves state-of-the-art performance and additional unlabeled data could further boost the captioning performance. Analysis of captions generated by our model shows that the generated captions are more discriminative and achieve higher self-retrieval performance than conventional methods.

\section{Related Work}

Image captioning methods can be divided into three categories~\cite{yao2017boosting}. \textbf{Template-based methods}~\cite{kulkarni2013babytalk,mitchell2012midge,yang2011corpus} generate captions based on language templates. \textbf{Search-based methods}~\cite{devlin2015language,farhadi2010every} search for the most semantically similar captions from a sentence pool. 
Recent works mainly focus on \textbf{language-based methods} with an encoder-decoder framework~\cite{vinyals2015show,mao2014deep,xu2015show,chen2015mind,wang2017skeleton,gan2017semantic,gu2017empirical,wu2016value,karpathy2015deep,gu2017stack}, where a convolutional neural network (CNN) encodes images into visual features, and an Long Short Term Memory network (LSTM) decodes features into sentences~\cite{vinyals2015show}. 
It has been shown that attention mechanisms~\cite{xu2015show,lu2017knowing,pedersoli2017areas,chen2017sca} and high-level attributes and concepts~\cite{you2016image,gan2017semantic,gu2017empirical,yao2017boosting} can help with image captioning. 
 
Maximum Likelihood Estimation (MLE) was adopted for training by many previous works. It maximizes the conditional likelihood of the next word conditioned on previous words. However, it leads to the exposure bias problem~\cite{ranzato2015sequence}, and the training objective does not match evaluation metrics. Training image captioning models by reinforcement learning techniques~\cite{sutton1998reinforcement} solves those problems~\cite{rennie2017self,ren2017deep,liu2017improved} and significantly improves captioning performance.

A problem of current image captioning models is that they tend to replicate phrases and sentences seen in the training set, and most generated captions follow certain templated patterns. Many recent works aim at increasing diversity of generated captions~\cite{vijayakumar2016diverse,wang2017diverse,wang2016diverse}. Generative adversarial networks (GAN) can be incorporated into captioning models to generate diverse and human-like captions~\cite{dai2017towards,shetty2017speaking}. Dai \etal~\cite{dai2017contrastive} proposed a contrastive learning technique to generate distinctive captions while maintaining the overall quality of generated captions. Vedantam \etal~\cite{vedantam2017context} introduced an inference technique to produce discriminative context-aware image captions using generic context-agnostic training data, but with a different problem setting from ours. It requires context information, \ie, a distractor class or a distractor image, for inference, which is not easy to obtain in generic image captioning applications.

In this work, we improve discriminativeness of captions by using a self-retrieval module to explicitly encourage generating discriminative captions during training. Based on the intuition that a discriminative caption should be able to successfully retrieve back the image corresponding to itself, the self-retrieval module performs text-to-image retrieval with the generated captions, serving as an evaluator of the captioning module. The retrieval reward for generated captions is back-propagated by REINFORCE algorithm. Our model can also be trained with partially labeled data to boost the performance. A concurrent work~\cite{luo2018discriminability} by Luo \etal also uses a discriminability objective similar to that of ours to generate discriminative captions. However, our work differs from it in utilizing unlabeled image data and mining moderately hard negative samples to further encourage discriminative captions.

\section{Methodology}

\begin{figure}[t]
\centering
\includegraphics[width=1\linewidth]{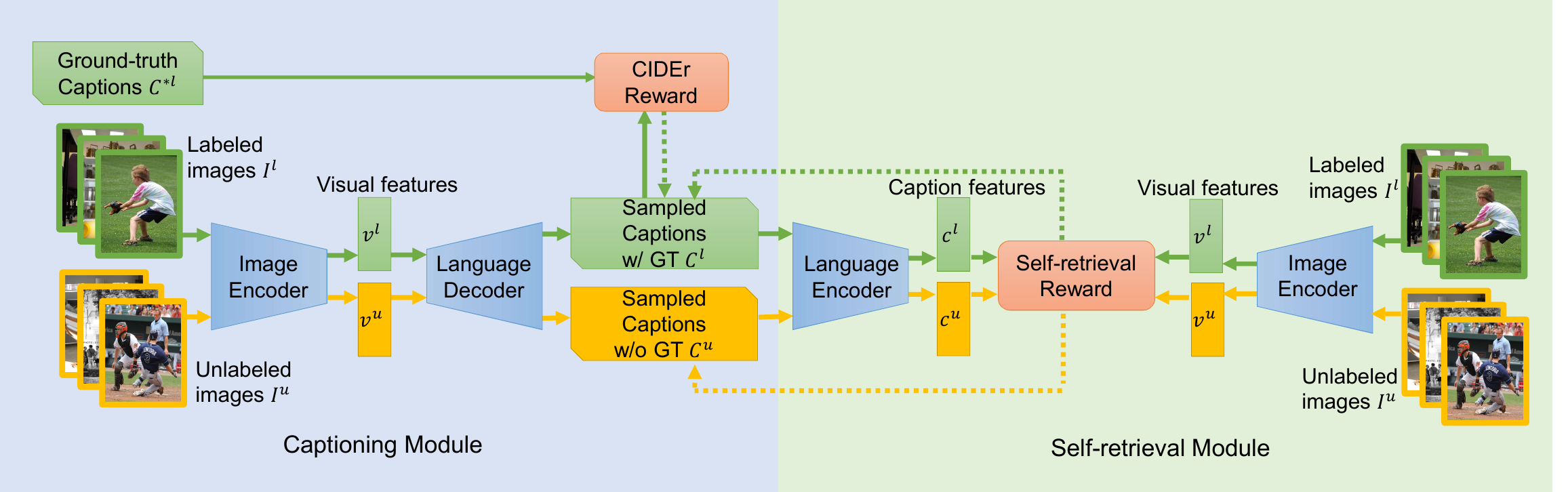}
\caption{Overall Framework of our proposed method. The captioning module (left) and the self-retrieval module (right) shares the same image encoder. Dotted lines mean that the reward for each sampled caption is back-propagated by REINFORCE algorithm. The captioning performance is improved by training the captioning module with text-to-image self-retrieval reward. Unlabeled images are naturally handled by our framework.}
\label{fig:framework}
\end{figure}

Given an image $I$, the goal of image captioning is to generate a caption $C = \{w_1,w_2,\dots,w_T\}$, where $w_i$ denotes the $i$th word, and we denote the ground-truth captions by $C^*=\{w^*_1,w^*_2,\dots,w^*_T\}$.

The overall framework, as shown in Fig.~\ref{fig:framework}, comprises of a captioning module and a self-retrieval module. 
The captioning module generates captions for given images. A Convolutional Neural Network (CNN) encodes images to visual features, and then a Long Short Term Memory network (LSTM) decodes a sequence of words based on the visual features.
The self-retrieval module is our key contribution, which is able to boost the performance of the captioning module with only partially labeled images. It first evaluates the similarities between generated captions with their corresponding input images and other distractor images. If the captioning module is able to generate discriminative enough descriptions, the similarity between the corresponding generated-caption-image pairs should be higher than those of non-corresponding pairs. Such constraint is modeled as a text-to-image retrieval loss and is back-propagated to the improve the captioning module by REINFORCE algorithm.


\subsection{Image Captioning with Self-retrieval Reward}\label{self_retrieval_reward}

\subsubsection{Captioning module.}\label{captioning_module}
The captioning module, aiming at generating captions for given images, is composed of a CNN image encoder $E_i(I)$ and an LSTM language decoder $D_c(v)$.
The image encoder $E_{i}$ encodes an image $I$ to obtain its visual features $v$, and the language decoder $D_{c}$ decodes the visual features $v$ to generate a caption $C$ that describes the contents of the image,
\begin{equation}
    v = E_{i}(I),~~ C = D_{c}(v).
\end{equation}

For conventional training by maximum-likelihood estimation (MLE), given the ground-truth caption words up to time step $t-1$, $\{w^*_1,\dots,w^*_{t-1}\}$, the model is trained to maximize the likelihood of $w^*_t$, the ground-truth word of time step $t$. Specifically, the LSTM outputs probability distribution of the word at time step $t$, given the visual features and ground-truth words up to time step $t-1$, and is optimized with the cross-entropy loss,
\begin{equation}
    L_{CE}(\theta)=-\sum^T_{t=1}\log (p_\theta(w^*_t|v,w^*_1,\dots,w^*_{t-1})),
\end{equation}
where $\theta$ represents learnable weights of the captioning model. 

For inference, since the ground-truth captions are not available, the model outputs the distribution of each word conditioned on previous generated words and visual features, $p_\theta(w_t|v,w_1,\dots,w_{t-1})$. The word at each time step $t$ is chosen based on the probability distribution of each word by greedy decoding or beam search.

\subsubsection{Self-retrieval module.}

A captioning model trained by MLE training often tends to imitate the word-by-word patterns in the training set. A common problem of conventional captioning models is that many captions are templated and generic descriptions (\eg ``A woman is standing on a beach''). 
Reinforcement learning with evaluation metrics (such as CIDEr) as reward~\cite{rennie2017self,liu2017improved} allows the captioning model to explore more possibilities in the sample space and gives a better supervision signal compared to MLE. However, the constraint that different images should not generate the same generic captions is still not taken into account explicitly. Intuitively, a good caption with rich details, such as ``A woman in a blue dress is walking on the beach with a black dog'', should be able to distinguish the corresponding image in context of other distractor images. To encourage such discriminative captions, we introduce the self-retrieval module to enforce the constraint that the generated captions should match its corresponding images better than other images. 

We therefore model the self-retrieval module to conduct text-to-image retrieval with the generated caption as a query. 
Since retrieving images from the whole dataset for each generated caption is time-consuming and infeasible during each training iteration, we consider text-to-image matching in each mini-batch. We first encode images and captions into features in the same embedding space a CNN encoder $E_{i}$ and a Gated Recurrent Unit (GRU) encoder $E_{c}$ for captions,
\begin{equation}
    v = E_{i}(I),~~ c = E_{c}(C),
\end{equation}
where $I$ and $C$ denote images and captions, and $v$ and $c$ denote visual features and caption features, respectively. 
Then the similarities between the embedded image features and caption features are calculated. The similarities between the features of a caption $c_i$ and the features of the $j$th image $v_j$ is denoted as $s(c_i,v_j)$. For a mini-batch of images $\{I_1,I_2,\cdots,I_n\}$ and a generated caption $C_i$ of the $i$th image, we adopt the triplet ranking loss with hardest negatives (\textit{VSE++}~\cite{faghri2017vse++}) for text-to-image retrieval,
\begin{equation}\label{vse++}
    L_{ret}(C_i,\{I_1,I_2,\cdots,I_n\})=\max_{j\neq i} [m-s(c_i,v_i)+s(c_i,v_j)]_+,
\end{equation}
where $[x]_+=\max (x,0)$. For a query caption $C_i$, we compare the similarity between the positive feature pair $\{c_i,v_i\}$ with the negative pairs $\{c_i,v_j\}$, where $j\neq i$. This loss forces the similarity of the positive pair to be higher than the similarity of the hardest negative pair by a margin $m$.
We also explore other retrieval loss formulations in Sec.~\ref{sec:ablation}.

The self-retrieval module acts as a discriminativeness evaluator of the captioning module, which encourages a caption generated from a given image by the captioning module to be the best matching to the given image among a batch of distractor images.

\subsubsection{Back-propagation by REINFORCE algorithm.}

For each input image, since self-retrieval is performed based on the complete generated caption, and sampling a word from a probability distribution is non-differentiable, we cannot back-propagate the self-retrieval loss to the captioning module directly. Therefore, REINFORCE algorithm is adopted to back-propagate the self-retrieval loss to the captioning module.

For image captioning with reinforcement learning, the LSTM acts as an ``agent'', and the previous generated words and image features are ``environment''. The parameters $\theta$ define the policy $p_\theta$ and at each step the model chooses an ``action'', which is the prediction of the next word based on the policy and the environment. Denote $C^s=\{w_1^s,\dots,w_T^s\}$ as the caption sampled from the predicted word distribution. Each sampled sentence receives a ``reward'' $r(C^s)$, which indicates its quality. Mathematically, the goal of training is to minimize the negative expected reward of the sampled captions,
\begin{equation}
  L_{RL}(\theta)=-\mathbb{E}_{C^s\sim p_\theta}[r(C^s)].
\end{equation}
Since calculating the expectation of reward over the policy distribution is intractable, we estimate it by Monte Carlo sampling based on the policy $p_{\theta}$. To avoid differentiating $r(C^s)$ with respect to $\theta$, we calculate the gradient of the expected reward by REINFORCE algorithm~\cite{williams1992simple},
\begin{equation}
    \nabla_\theta L_{RL}(\theta)=-\mathbb{E}_{C^s\sim p_\theta}[r(C^s)\nabla_{\theta}\log p_\theta(C^s)].
\end{equation}
To reduce the variance of the gradient estimation, we subtract the reward with a baseline $b$, without changing the expected gradient~\cite{sutton1998reinforcement}. $b$ is chosen as the reward of greedy decoding captions~\cite{rennie2017self}.
\begin{equation}
  \nabla_\theta L_{RL}(\theta)=-\mathbb{E}_{C^s\sim p_\theta}[(r(C^s)-b)\nabla_{\theta}\log p_\theta(C^s)].
\end{equation}
For calculation simplicity, the expectation is approximated by a single Monte-Carlo sample from $p_\theta$,
\begin{equation}
  \nabla_\theta L_{RL}(\theta)\approx -(r(C^s)-b)\nabla_{\theta}\log p_\theta(C^s).
\end{equation}

In our model, for each sampled caption $C^s_i$, we formulate the reward as a weighted summation of its CIDEr score and the self-retrieval reward, which is the negative caption-to-image retrieval loss.
\begin{equation}
    r(C^s_i)=r_{cider}(C^s_i) + \alpha \cdot r_{ret}(C^s_i,\{I_1,\cdots,I_n\}),
\end{equation}
where $r_{cider}(C^s_i)$ denotes the CIDEr score of $C^s_i$, $r_{ret}=-L_{ret}$ is the self-retrieval reward, and $\alpha$ is the weight to balance the rewards. 
The CIDEr reward ensures that the generated captions are similar to the annotations, and the self-retrieval reward encourages the captions to be discriminative. 
By introducing this reward function, we can optimize the sentence-level reward through sampled captions.

\subsection{Improving Captioning with Partially Labeled Images}\label{partially_labeled}

\subsubsection{Training with partially labeled data.}

The self-retrieval module compares a generated caption with its corresponding image and other distractor images in the mini-batch. As the caption-image correspondence is incorporated naturally in caption generation, \ie, a caption with the image it is generated from automatically form a positive caption-image pair, and with other images form negative pairs, our proposed self-retrieval reward does not require ground-truth captions. So our framework can generalize to semi-supervised setting, where a portion of images do not have ground-truth captions. Thus more training data can be involved in training without extra annotations.

We mix labeled data and unlabeled data with a fixed proportion in each mini-batch. Denote the labeled images in a mini-batch as $\{I^l_1,I^l_2,\cdots,I^l_{n_l}\}$, and their generated captions as $\{C^l_1,C^l_2,\cdots,C^l_{n_l}\}$. Denote unlabeled images in the same mini-batch as $\{I^u_1,I^u_2,\cdots,I^u_{n_u}\}$ and the corresponding generated captions as $\{C^u_1,C^u_2,\cdots,C^u_{n_u}\}$. The reward for labeled data is the composed of the CIDEr reward and self-retrieval reward computed in the mini-batch for each generated caption,
\begin{equation}\label{equ:labeled_reward}
    r(C^l_i)=r_{cider}(C^l_i) + \alpha \cdot r_{ret}(C^l_i,\{I^l_1,\cdots,I^l_{n_l}\}\cup\{I^u_1,\cdots,I^u_{n_u}\}).
\end{equation}
The retrieval reward $r_{ret}$ compares the similarity between a caption and the corresponding image, with those of all other labeled and unlabeled images in the mini-batch, to reflect how well the generated caption can discriminate its corresponding image from other distractor images.

As CIDEr reward cannot be computed without ground-truth captions, the reward for unlabeled data is only the retrieval reward computed in the mini-batch,
\begin{equation}\label{equ:unlabeled_reward}
    r(C^u_i) = \alpha \cdot r_{ret}(C^u_i,\{I^l_1,\cdots,I^l_{n_l}\}\cup\{I^u_1,\cdots,I^u_{n_u}\}).
\end{equation}
In this way, the unlabeled data could also be used in training without captioning annotations, to further boost the captioning performance.

\subsubsection{Moderately Hard Negative Mining in Unlabeled Images.}

\begin{figure}[t]
\centering
\includegraphics[width=1\linewidth]{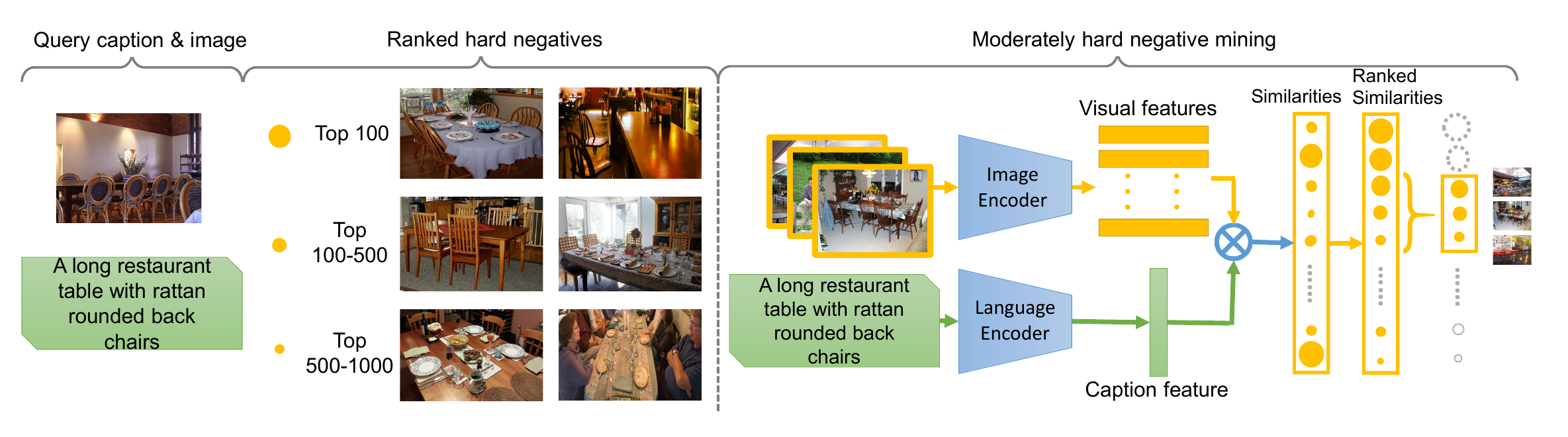}
\caption{Moderately hard negative mining. The left part shows a ground-truth caption and its top hard negatives mined from unlabeled images. The right part shows the process of moderately hard negative mining. The circles of different sizes stand for the similarities between each image and the query caption.}
\label{hard_mining}
\end{figure}

As described before, the self-retrieval reward is calculated based on the similarity between positive (corresponding) caption-image pairs and negative (non-corresponding) pairs. The training goal is to maximize the similarities of positive pairs and minimize those of negative pairs. To further encourage discriminative captions, we introduce hard negative caption-image pairs in each mini-batch. For example, in Fig.~\ref{fig1_current_captions}, although the first two images are similar, humans are not likely to describe them in the same way. We would like to encourage captions that can distinguish the second image from the first one (\eg, ``Creative centerpiece floral arrangement at an outdoor table''), instead of a generic description (\eg, ``A vase sitting on a table'').

However, an important observation is that choosing the hardest negatives may impede training. This is because images and captions do not always follow strictly one-to-one mapping. In the left part of Fig.~\ref{hard_mining}, we show a ground-truth caption and its hard negatives mined from unlabeled images. The top negative images from the unlabeled dataset often match well with the ground-truth captions from the labeled dataset. For example, when the query caption is ``A long restaurant table with rattan rounded back chairs'', some of the retrieved top images can also match the caption well. So directly taking the hardest negative pairs is not optimal. Therefore, we propose to use \emph{moderately hard negatives} of the generated captions instead of the hardest negatives.

We show moderately hard negative mining in the right part of Fig.~\ref{hard_mining}. We encode a ground-truth caption $C^*$ from the labeled dataset into features $c^*$ and all unlabeled images $\{I^u_1,\cdots,I^u_{n_u}\}$ into features $\{v^u_1,\cdots,v^u_{n_u}\}$. The similarities $\{s(c^*,v^u_1),\cdots,s(c^*,v^u_{n_u})\}$ between the caption and each unlabeled image are derived by the retrieval model.
Then we rank the unlabeled images by the similarities between each image and the query caption $C^*$ in a descending order. 
Then the indexes of moderately hard negatives are randomly sampled from a given range $[h_{min},h_{max}]$. The sampled hard negatives from unlabeled images and the captions' corresponding images from the labeled dataset together form a mini-batch.

By moderately hard negative mining, we select proper samples for training, encouraging the captioning model to generate captions that could discriminate the corresponding image from other distractor images.

\subsection{Training Strategy}\label{training_strategy}

We first train the text-to-image self-retrieval module with all training images and corresponding captions in the labeled dataset.
The captioning module shares the image encoder with the self-retrieval module. When training the captioning module, the retrieval module and CNN image encoder are fixed. 

For captioning module, we first pre-train it with cross-entropy loss, to provide a stable initial point, and reduce the sample space for reinforcement learning.
The captioning module is then trained by REINFORCE algorithm with CIDEr reward and self-retrieval reward with either fully labeled data or partially labeled data. The CIDEr reward guarantees the generated captions to be similar to ground-truth captions, while the self-retrieval reward encourages the generated captions to be discriminative. 
For labeled data, the reward is the weighted sum of CIDEr reward and self-retrieval reward (Eq.~(\ref{equ:labeled_reward})), and for unlabeled data, the reward is only the self-retrieval reward (Eq.~(\ref{equ:unlabeled_reward})). The unlabeled data in each mini-batch is chosen by moderately hard negative mining from unlabeled data.
Implementation details can be found in Sec.~\ref{imp_detail}.

\section{Experiments}
\subsection{Datasets and Evaluation Criteria}

We perform experiments on COCO and Flickr$30$k captioning datasets. For fair comparison, we adopt the widely used Karpathy split~\cite{karpathy2015deep} for COCO dataset, which uses 5,000 images for validation, 5,000 for testing, and the rest 82,783 for training.
For data preprocessing, we first convert all characters into lower case and remove the punctuations. Then we replace words that occur less than 6 times with an `UNK' token. The captions are truncated to be no more than 16 words during training.
When training with partially labeled data, we use the officially released COCO unlabeled images as additional data without annotations.
The widely used BLEU~\cite{papineni2002bleu}, METEOR~\cite{denkowski2014meteor}, ROUGE-L~\cite{lin2004rouge}, CIDEr-D~\cite{vedantam2015cider} and SPICE~\cite{spice2016} scores are adopted for evaluation.

\subsection{Implementation Details}\label{imp_detail}
 
\noindent \textbf{Self-retrieval module.}
For the self-retrieval module, each word is embedded into a 300-dimensional vector and inputted to the GRU language encoder, which encodes a sentence into 1024-dimensional features. The image encoder is a ResNet-101 model, which encodes an image into 2048-dimensional visual features. Both the encoded image features and sentence features are projected to the joint embedding space of dimension 1024. The similarity between image features and sentence features is the inner product between the normalized feature vectors. We follow the training strategy in~\cite{faghri2017vse++}. 

\noindent \textbf{Captioning module.}
The captioning module shares the same image encoder with the self-retrieval module. The self-retrieval module and image encoder are fixed when training the captioning module. We take the $2048\times7\times7$ features before the average pooling layer from ResNet-101 as the visual features. For the language decoder, we adopt a topdown attention LSTM and a language LSTM, following the Top-Down attention model in~\cite{anderson2017bottom}. We do not use Up-Down model in the same paper, because it involves an object detection model and requires external data and annotations from Visual Genome~\cite{krishna2017visual} for training.

The captioning module is trained with Adam~\cite{kingma2014adam} optimizer. 
The model is first pre-trained by cross-entropy loss, and then trained by REINFORCE. 
Restart technique~\cite{loshchilov2016sgdr} is used improve the model convergence. 
We use scheduled sampling~\cite{bengio2015scheduled} and increase the probability of feeding back a sample of the word posterior by 0.05 every 5 epochs, until the feedback probability reaches 0.25. 
We set the weight of self-retrieval reward $\alpha$ to 1. For training with partially labeled data, the proportion of labeled and unlabeled images in a mini-batch is 1:1.

\noindent \textbf{Inference.}
For inference, we use beam search with beam size 5 to generate captions. Specifically, we select the top 5 sentences with the highest probability at each time step, and consider them as the candidates based on which to generate the next word. We do not use model ensemble in our experiments.

\subsection{Results}

\textbf{Quantitative results.}
We compare our captioning model performance with existing methods on COCO and Flickr$30$k datasets in Table~\ref{result_coco} and Table~\ref{result_f30k}. The models are all pre-trained by cross-entropy loss and then trained with REINFORCE algorithm. The baseline model is the captioning module only trained with only CIDEr reward. The SR-FL model is our proposed framework trained with fully labeled data, with both CIDEr and self-retrieval rewards. The SR-PL model is our framework trained with partially labeled data (all labeled data and additional unlabeled data), with both rewards for labeled images and only self-retrieval reward for unlabeled images. 
It is shown from the results that the baseline model without self-retrieval module is already a strong baseline. Incorporating the self-retrieval module with fully-labeled data (SR-FL) improves most metrics by large margins. Training with additional unlabeled data (SR-PL) further enhances the performance. The results validate that discriminativeness is crucial to caption quality, and enforcing this constraint by self-retrieval module leads to better captions. 

\begin{table}[tb]
\caption{Single-model performance by our proposed method and state-of-the-art methods on COCO standard Karpathy test split.}
\centering
\scriptsize
\begin{tabular}{c|cccccccc}
\hline
Methods        & CIDEr & SPICE & BLEU-1 & BLEU-2 & BLEU-3 & BLEU-4 & METEOR & ROUGE-L \\ \hline
Hard-attention~\cite{xu2015show} & -     & -     & 71.8   & 50.4   & 35.7   & 25.0   & 23.0   & -       \\
Soft-attention~\cite{xu2015show} & -     & -     & 70.7   & 49.2   & 34.4   & 24.3   & 23.9   & -       \\
VAE~\cite{pu2016variational}           & 90.0  & -     & 72.0   & 52.0   & 37.0   & 28.0   & 24.0   & -       \\
ATT-FCN~\cite{you2016image}        & -     & -     & 70.9   & 53.7   & 40.2   & 30.4   & 24.3   & -       \\
Att-CNN+RNN~\cite{wu2016value}    & 94.0  & -     & 74.0   & 56.0   & 42.0   & 31.0   & 26.0   & -       \\
SCN-LSTM~\cite{gan2017semantic}       & 101.2 & -     & 72.8   & 56.6   & 43.3   & 33.0   & 25.7   & -       \\
Adaptive~\cite{lu2017knowing}       & 108.5 & -     & 74.2   & 58.0   & 43.9   & 33.2   & 26.6   & -       \\
SCA-CNN~\cite{chen2017sca}        & 95.2  & -     & 71.9   & 54.8   & 41.1   & 31.1   & 25.0   & 53.1    \\
SCST-Att2all~\cite{rennie2017self}   & 114.0 & -     & -      & -      & -      & 34.2   & 26.7   & 55.7    \\
LSTM-A~\cite{yao2017boosting}         & 100.2 & 18.6  & 73.4   & 56.7   & 43.0   & 32.6   & 25.4   & 54.0    \\
DRL~\cite{ren2017deep}            & 93.7  & -     & 71.3   & 53.9   & 40.3   & 30.4   & 25.1   & 52.5    \\
Skeleton Key~\cite{wang2017skeleton}   & 106.9 & -     & 74.2   & 57.7   & 44.0   & 33.6   & 26.8   & 55.2    \\
CNNL+RHN~\cite{gu2017empirical}       & 98.9  & -     & 72.3   & 55.3   & 41.3   & 30.6   & 25.2   & -       \\
TD-M-ATT~\cite{chen2017temporal}       & 111.6 & -     & 76.5   & 60.3   & 45.6   & 34.0   & 26.3   & 55.5    \\ 
ATTN+C+D(1)~\cite{luo2018discriminability} & 114.25 & \textbf{21.05} & - & - & - & \textbf{36.14} & 27.38 & \textbf{57.29} \\ \hline
Ours-baseline  & 112.7 & 20.0  & 79.7   & 62.2   & 47.1   & 35.0   & 26.7   & 56.4    \\
Ours-SR-FL     & 114.6 & 20.5  & 79.8   & 62.3   & 47.1   & 34.9   & 27.1   & 56.6    \\
Ours-SR-PL     & \textbf{117.1} & 21.0  & \textbf{80.1}   & \textbf{63.1}   & \textbf{48.0}   & 35.8   & \textbf{27.4}   & 57.0    \\ \hline
\end{tabular}
\label{result_coco}
\end{table}
\begin{table}[]
\caption{Single-model performance by our proposed method and state-of-the-art methods on Flickr$30$k.}
\scriptsize
\centering
\begin{tabular}{c|cccccccc}
\hline
Methods        & CIDEr & SPICE & BLEU-1 & BLEU-2 & BLEU-3 & BLEU-4 & METEOR & ROUGE-L \\ \hline
Hard-attention~\cite{xu2015show} & -     & -     & 66.9   & 43.9   & 29.6   & 19.9   & 18.5   & -       \\
Soft-attention~\cite{xu2015show} & -     & -     & 66.7   & 43.4   & 28.8   & 19.1   & 18.5   & -       \\
VAE~\cite{pu2016variational}            & -     & -     & 72.0   & 53.0   & 38.0   & 25.0   & -      & -       \\
ATT-FCN~\cite{you2016image}        & -     & -     & 64.7   & 46.0   & 32.4   & 23.0   & 18.9   & -       \\
Att-CNN+RNN~\cite{wu2016value}    & -     & -     & 73.0   & 55.0   & 40.0   & 28.0   & -      & -       \\
SCN-LSTM~\cite{gan2017semantic}       & -     & -     & 73.5   & 53.0   & 37.7   & 25.7   & 21.0   & -       \\
Adaptive~\cite{lu2017knowing}       & 53.1  &       & 67.7   & 49.4   & 35.4   & 25.1   & 20.4   & -       \\
SCA-CNN~\cite{chen2017sca}        & -     & -     & 66.2   & 46.8   & 32.5   & 22.3   & 19.5   & -       \\
CNNL+RHN~\cite{gu2017empirical}       & 61.8  & 15.0  & \textbf{73.8}   & \textbf{56.3}   & \textbf{41.9}   & \textbf{30.7}   & 21.6   & -       \\ \hline
Ours-baseline  & 57.1  & 14.2  & 72.8   & 53.4   & 38.0   & 27.1   & 20.7   & 48.5    \\
Ours-SR-FL     & 61.7  & 15.3  & 72.0   & 53.4   & 38.5   & 27.8   & 21.5   & 49.4    \\
Ours-SR-PL     & \textbf{65.0}  & \textbf{15.8}  & 72.9   & 54.5   & 40.1   & 29.3   & \textbf{21.8}   & \textbf{49.9}    \\ \hline
\end{tabular}
\label{result_f30k}
\end{table}

\begin{figure}[t]
\centering
\includegraphics[width=1\linewidth]{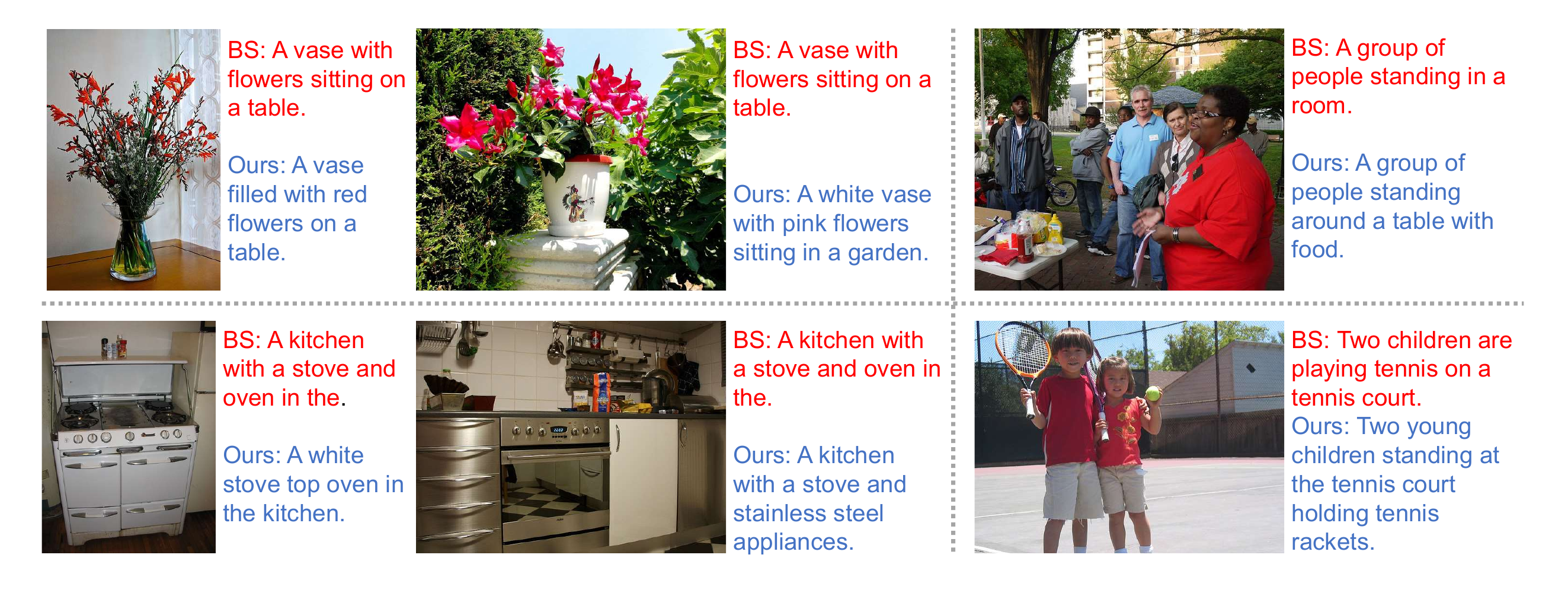}
\caption{Qualitative results by baseline model and our proposed model.}
\label{results}
\end{figure}
\noindent \textbf{Qualitative results.} 
Fig.~\ref{results} shows some examples of our generated captions and ground-truth captions. Both the baseline model and our model with self-retrieval reward can generate captions relevant to the images. However, it is easy to observe that our model can generate more discriminative captions, while the baseline model generates generic and templated captions. For example, the first and the second images in the first row share slightly different contents. The baseline model fails to describe their differences and generates identical captions ``A vase with flowers sitting on a table''. But our model captures the distinction, and expresses with sufficient descriptive details ``red flowers'', ``white vase'' and ``in a garden'' that help to distinguish those images. The captions for the last images in both rows show that the baseline model falls into a stereotype and generates templated captions, because of a large number of similar phrases in the training set. However, captions generated by our model alleviates this problem, and generates accurate descriptions for the images.

\subsection{Ablation Study}\label{sec:ablation}
\begin{table}[tb]
\caption{Ablation study results on COCO.}
\centering
\scriptsize
\begin{tabular}{c|c|cccccc}
\hline
\multicolumn{2}{c|}{Experiment Settings}                                                                     & CIDEr & SPICE & BLEU-3 & BLEU-4 & METEOR & ROUGE-L \\ \hline
\multicolumn{2}{c|}{Baseline}                                                                                & 112.7 & 20.0  & 47.1   & 35.0   & 26.7   & 56.4    \\ \hline
\multirow{3}{*}{Retrieval Loss}                                                               & VSE++        & \textbf{117.1} & \textbf{21.0}  & \textbf{48.0}   & \textbf{35.8}   & \textbf{27.4}   & \textbf{57.0}    \\
                                                                                              & VSE0         & 116.9 & 20.9  & 47.7   & 35.7   & \textbf{27.4}   & 56.8    \\
                                                                                              & softmax      & 114.5 & 20.5  & 46.8   & 34.6   & 27.1   & 56.5    \\ \hline
\multirow{3}{*}{\begin{tabular}[c]{@{}c@{}}Weight of \\ Self-retrieval\\ Reward $\alpha$\end{tabular}} & 0           & 112.7  & 20.0  & 47.1   & 35.0   & 26.7   & 56.4    \\
                                                                                              & 1            & \textbf{117.1} & \textbf{21.0}  & \textbf{48.0}   & \textbf{35.8}   & \textbf{27.4}   & \textbf{57.0}    \\
                                                                                              & 4            & 113.7 & 20.5  & 46.5   & 34.3   & 27.0   & 56.5    \\ \hline
\multirow{3}{*}{\begin{tabular}[c]{@{}c@{}}\ Ratio between labeled\\ \ and unlabeled\end{tabular}} & 1:2          & 115.4 & 20.5  & 46.8   & 34.7   & 27.2   & 56.6    \\
                                                                                              & 1:1          & \textbf{117.1} & \textbf{21.0}  & \textbf{48.0}   & \textbf{35.8}   & \textbf{27.4}   & \textbf{57.0}    \\
                                                                                              & 2:1          & 115.0 & 20.5  & 46.8   & 34.7   & 27.2   & 56.7    \\ \hline
\multirow{3}{*}{\begin{tabular}[c]{@{}c@{}}Hard Negative\\ Index Range\end{tabular}}          & no hard mining & 114.6 & 20.7  & 46.7   & 34.6   & 27.3   & 56.7    \\ 
                                                                                              & top 100      & 114.1 & 20.3  & 46.6   & 34.5   & 27.0   & 56.4    \\
                                                                                              & top 100-1000 & \textbf{117.1} & \textbf{21.0}  & \textbf{48.0}   & \textbf{35.8}   & \textbf{27.4}   & \textbf{57.0}    \\    \hline
                                                                                              
\end{tabular}
\label{ablation}
\end{table}
\textbf{Formulation of self-retrieval loss.}
As described in Sec.~\ref{self_retrieval_reward}, the self-retrieval module requires a self-retrieval loss to measure the discriminativeness of the generated captions. Besides \textit{VSE++} loss (Eq.~(\ref{vse++})), we explore triplet ranking loss without hard negatives, denoted by \textit{VSE0},
\begin{equation}\label{vse0}
    L_{ret}(C_i,\{I_1,I_2,\cdots,I_n\})=\sum_{j\neq i} [m-s(c_i,v_i)+s(c_i,v_j)]_+,
\end{equation}
and softmax classification loss, denoted by \textit{softmax},
\begin{equation}\label{softmax}
    L_{ret}(C_i,\{I_1,I_2,\cdots,I_n\})=-\log \frac{\exp\left(s(c_i ,v_i)/T\right)}{\sum_{j=1}^n \exp \left(s(c_i,v_j)/T\right)},
\end{equation}
where $T$ is the temperature parameter that normalizes the caption-image similarity to a proper range.
We show the results trained by the three loss formulations in Table~\ref{ablation}.\footnote{For the reported results in all experiments and ablation study, we tuned hyper-parameters on the validation set and directly used validation’s best point to report results on the test set.} All of those loss formulations lead to better performance compared to the baseline model, demonstrating the effectiveness of our proposed self-retrieval module. Among them, \textit{vse++} loss performs slightly better, which is consistent with the conclusion in~\cite{faghri2017vse++} that \textit{vse++} loss leads to better visual-semantic embeddings.

\noindent \textbf{Balance between self-retrieval reward and CIDEr reward.}
During training by REINFORCE algorithm, the total reward is formulated as the weighted summation of CIDEr reward and self-retrieval reward, as shown in Eq.~(\ref{equ:labeled_reward}). To determine how much each of them should contribute to training, we investigate how the weight between them should be set. As shown in Table~\ref{ablation}, we investigate $\{0, 1, 4\}$ for the weight of self-retrieval reward $\alpha$, and the results indicate that $\alpha=1$ leads to the best performance. Too much emphasis on self-retrieval reward will harm the model performance, because it fails to optimize the evaluation metric CIDEr. This shows that both CIDEr and our proposed self-retrieval reward are crucial and their contributions need to be balanced properly.

\noindent \textbf{Proportion of labeled and unlabeled data.}
When training with partially labeled data, we use a fixed proportion between labeled and unlabeled images. We experiment on the proportion of forming a mini-batch with labeled and unlabeled data. We try three proportions, 1:2, 1:1 and 2:1, with the same self-retrieval reward weight $\alpha=1$. The results in Table~\ref{ablation} show that the proportion of 1:1 leads to the best performance.

\noindent \textbf{Moderately Hard Negative Mining.}
In Sec.~\ref{partially_labeled}, we introduce how to mine semantically similar images from unlabeled data to provide moderately hard negatives for training. We analyze the contribution of moderately hard negative mining in Table~\ref{ablation}. Firstly, the performance gain is relatively low without hard negative mining, demonstrating the effectiveness of this operation. Secondly, after ranking unlabeled images based on the similarity between the given ground-truth caption and unlabeled images in the descending order, the index range $[h_{min},h_{max}]$ of selecting hard negatives also impacts results. 
There are cases that an unlabeled image is very similar to an image in the training set, and a caption may naturally correspond to several images. Therefore, selecting the hardest negatives is very likely to confuse the model. 
In our experiments, we found that setting the hard negative index range $[h_{min},h_{max}]$ to [100, 1000] for the ranked unlabeled images is optimal.

\subsection{Discriminativeness of Generated Captions}

\begin{table}[tb]
\caption{Text-to-image retrieval performance, and uniqueness and novelty of generated captions by different methods on COCO.}
\centering
\scriptsize
\label{my-label}
\begin{tabular}{c|ccc|cc}
\hline
\multirow{2}{*}{Methods} & \multicolumn{3}{c|}{Generated-caption-to-image retrieval} & \multicolumn{2}{c}{Uniqueness and novelty evaluation} \\ \cline{2-6} 
                         & ~~recall@1~~~         & ~recall@5~        & recall@10        & unique captions            & novel captions           \\ \hline
Skeleton Key~\cite{wang2017skeleton}              & -                  & -                 & -                & 66.96\%                    & 52.24\%                  \\ \hline标
Ours-baseline            & 27.5               & 59.3              & 74.0               & 61.56\%                    & 51.38\%                  \\
Ours-SR-PL               & \textbf{33.0}               & \textbf{66.4}              & \textbf{80.1}             & \textbf{72.34\%}                    & \textbf{61.52\%}                  \\ \hline
\end{tabular}
\label{discriminativeness}
\end{table}

\textbf{Retrieval performance by generated captions.}
Since the self-retrieval module encourages discriminative captions, we conduct an experiment to retrieve images with the generated captions as queries, to validate that captions generated by our model are indeed more discriminative than those generated by the model without self-retrieval module. Different from the self-retrieval study in~\cite{dai2017contrastive}, which uses the conditional probabilities of generated captions given images to obtain a ranked list of images, we perform self-retrieval by our self-retrieval module. More precisely, we rank the images based on the similarities between the images and a generated query sentence calculated by our retrieval module. We compute the recall of the corresponding image that appears in the top-$k$ ranked images. The retrieval performance is an indicator of the discriminativeness of generated captions. In Table~\ref{discriminativeness}, we report retrieval results on COCO Karpathy test split. It can be clearly seen that the our model improves the retrieval performance by a large margin.

\noindent \textbf{Uniqueness and novelty evaluation.}
A common problem for captioning models is that they have limited ability of generating captions that have not been seen in the training set, and generates identical sentences for similar images~\cite{devlin2015language}. This demonstrates that the language decoder is simply repeating the sequence patterns it observed in the training set. Although our approach is not directly designed to improve diversity or encourage novel captions, we argue that by encouraging discriminative captions, we can improve the model's ability to generate unique and novel captions. Following the measurements in~\cite{wang2017skeleton}, we evaluate the percentage of unique captions (captions that are unique in all generated captions) and novel captions (captions that have not been seen in training) on COCO Karpathy test split. It is shown in Table~\ref{discriminativeness} that our framework significantly improves uniqueness and novelty of the generated captions.

\section{Conclusions}

In this work, we address the problem that captions generated by conventional approaches tend to be templated and generic. We present a framework that explicitly improves discriminativeness of captions via training with self-retrieval reward. The framework is composed of a captioning module and a novel self-retrieval module, which boosts discriminativeness of generated captions. The self-retrieval reward is back-propagated to the captioning module by REINFORCE algorithm. Results show that we obtain more discriminative captions by this framework, and achieve state-of-the-art performance on two widely used image captioning datasets.

\section*{Acknowledgement}
This work is supported by SenseTime Group Limited, the General Research Fund sponsored by the Research Grants Council of Hong Kong (Nos. CUHK14213616, CUHK14206114, CUHK14205615, CUHK14203015, CUHK14239816, CUHK419412, CUHK14207814, CUHK14208417, CUHK14202217), the Hong Kong Innovation and Technology Support Program (No.ITS/121/15FX).


\bibliographystyle{splncs04}
\bibliography{ref}

\end{document}